\documentclass[11pt]{article}

\usepackage{acl}

\usepackage{times}
\usepackage{latexsym}
\usepackage{amsmath, amsfonts}
\usepackage[T1]{fontenc}

\usepackage[utf8]{inputenc}

\usepackage{microtype}
\usepackage{tikz}
\usepackage{multirow}
\usepackage{diagbox}
\usepackage{multicol}
\usepackage{caption}
\usepackage{graphicx}
\usepackage{subfigure}
\usepackage{booktabs}
\usepackage{stfloats}
\usepackage{colortbl}
\usepackage{cite}
\usepackage{setspace}

\usepackage[linesnumbered,ruled,vlined]{algorithm2e}
\usepackage{url}
\usepackage{hyperref}
\makeatletter
\def\UrlAlphabet{%
      \do\a\do\b\do\c\do\d\do\e\do\f\do\g\do\h\do\i\do\j%
      \do\k\do\l\do\m\do\n\do\o\do\p\do\q\do\r\do\s\do\t%
      \do\u\do\v\do\w\do\x\do\y\do\z\do\A\do\B\do\C\do\D%
      \do\E\do\F\do\G\do\H\do\I\do\J\do\K\do\L\do\M\do\N%
      \do\O\do\P\do\Q\do\R\do\S\do\T\do\U\do\V\do\W\do\X%
      \do\Y\do\Z}
\def\UrlDigits{\do\1\do\2\do\3\do\4\do\5\do\6\do\7\do\8\do\9\do\0}
\g@addto@macro{\UrlBreaks}{\UrlOrds}
\g@addto@macro{\UrlBreaks}{\UrlAlphabet}
\g@addto@macro{\UrlBreaks}{\UrlDigits}
 
\usepackage{amsmath,amssymb,amsthm,mathrsfs}
\usepackage{algorithmic}
\usepackage{textcomp}
\usepackage{xcolor}
\usepackage{color}
\newcommand{\yue}[1] {{\color{blue}#1}}

\title{Cybersecurity Entity Alignment via Masked Graph Attention Networks}

\author{
Yue Qin\\
Indiana University Bloomington\\
\texttt{qinyue@iu.edu} \\\And
Xiaojing Liao\\
Indiana University Bloomington\\
\texttt{xliao@indiana.edu}\\      
}

\newcommand{\ignore}[1]{}

\begin{document}
\maketitle
\begin{abstract}
Cybersecurity vulnerability information is often recorded by multiple channels, including government vulnerability repositories, individual-maintained vulnerability-gathering platforms, or vulnerability-disclosure email lists and forums. Integrating vulnerability information from different channels enables comprehensive threat assessment and quick deployment to various security mechanisms. Efforts to automatically gather such information, however, are impeded by the limitations of today's entity alignment techniques. In our study, we annotate the first cybersecurity-domain entity alignment dataset and reveal the unique characteristics of security entities. Based on these observations, we propose the first cybersecurity entity alignment model, CEAM
%
, which equips GNN-based entity alignment with two mechanisms: \textit{asymmetric masked aggregation} and \textit{partitioned attention}. Experimental results on cybersecurity-domain entity alignment datasets demonstrate that CEAM significantly outperforms state-of-the-art entity alignment methods.
\end{abstract}

\section{Introduction}

There are many channels of cybersecurity vulnerability reports: Public vulnerability databases, including prominent U.S. government vulnerability repositories (e.g., the National Vulnerability Database),  individual-maintained vulnerability-gathering platforms (e.g., Security Focus), vulnerability disclosure email lists and forums, and many others.
Integrating such vulnerability information is essential for an organization to gain a big picture of the fast-evolving vulnerability landscape, timely identify early signs of cybersecurity risk, and effectively contain the threat with proper means. 
However, it is non-trivial to link the same vulnerabilities among different sources, as unique identifiers (e.g., CVE-ID) may not be mentioned in some repositories. 
The key step of cybersecurity information integration is to link security entities (e.g., vulnerabilities)
that refer to the same real-world entity 
across different data sources.
This is an entity alignment (EA) problem on security Knowledge Graphs, which can be programmatically constructed from structured or semi-structured security reports.
%
%
However, to the best of our knowledge, no previous study has investigated domain-specific entity alignment in the field of cybersecurity.
%
%

Traditional entity alignment methods based on hand-crafted rules are less successful since different KGs vary in structures and textual features ~\citep{zeng2021comprehensive}.
More recent work~\citep{hu2019multike} utilize Knowledge Graph Embedding (KGE) models trained towards triple plausibility (e.g., TransE~\citep{bordes2013translating}) to align equivalent entities into a unified vector space based on a few seed alignments. 
However, such methods are not suitable for security entity alignment since the KGE models cannot generate embeddings for new entities added to the KG after training. The security information is always timely updated, and it is unrealistic to retrain the KG embedding model on the whole augmented graph each time new security entities (e.g., vulnerabilities) are revealed. 
 \citep{seu} regards entity alignment as an assignment problem between two isomorphic graphs by reordering the entity node indices. However, it's difficult to be adapted for security entity alignment because security KGs crafted from different repositories diverge in graph topology, and only a \textit{subset} of entities on both KGs can be aligned. 
Inspired by the recent success of Graph Neural Network (GNN) on open-domain entity alignment~\citep{wu2020neighborhood, rrea, emgcn}, we propose a cybersecurity domain-specific entity alignment model to recognize the equivalent cyber-vulnerabilities. 
The GNN mechanism enables recursive information propagation among neighbors to learn structure-aware entity representations for the alignment.
%
%
However, 
the core assumptions that identical entities have similar attributes and  neighbors 
 and vice versa 
 do not hold for cross-platform security entities. 
In our study, we observe identical vulnerability show inconsistent attributes (vulnerability artifacts, e.g, impact and affected version) in different vulnerability repositories (see Section \ref{sec:measurement}).
The main reason is that a vulnerability is sometimes assessed considering specific execution environments such as operating systems, architectures, configurations, and organization policies, which yields different vulnerability artifacts.
In addition, different repositories provide vulnerability artifacts in different granularity (e.g., level of details), especially for the vulnerabilities disclosed in maillists and forums.

\vspace{-3pt}
Therefore, in this paper, we propose the first cybersecurity-specific entity alignment model, CEAM. It equips GNN-based entity alignment model with two mechanisms: \textit{asymmetric masked aggregation} and \textit{partitioned attention}, to address the above challenges.
%
We first aggregate \textit{selective} attribute information to learn the semantic embeddings for security entities by an asymmetric mask. It propagates only a specific portion of attributes to the target entity, enforcing that the distance between two entities is narrowed down only when they are mutually the nearest candidate of each other. 
Further, we use GNNs to update entity embeddings with structural information based on graph topology, where the partitioned attention mechanism ensures that the artifacts critical to the vulnerability identification are always taken more consideration during the propagation.  
%
%
Finally, we use two-layer MLP (Mutilayer Perceptron) to decide whether two entities are identical according to the discrepancy between entity embeddings learned by GNNs.
Our experiments show that the two mechanisms
collectively improve alignment quality by 10.3\% F1 score in average. Overall, the model achieves 81.5\% F1 score.
%
%
%
The contributions of this paper are as follows: 

\noindent$\bullet$ We proposed the \textit{first} cybersecurity-domain entity alignment model infused with domain knowledge towards cybersecurity information aggregation. 

\noindent$\bullet$ We fixed the severe information loss (i.e., missing vulnerability identifiers) from the alignment results in even high-profile sources which deliver the most informative vulnerability presentation.


\noindent$\bullet$ We release the \textit{first} annotated security KGs and cybersecurity-domain entity alignment datasets.

\section{Background}

\subsection{Vulnerability profiling standard}
\label{sec:background-vul}
According to CNA\footnote{\begin{scriptsize}
CVE Numbering Authorities: https://cve.mitre.org/cve/cna.html
\end{scriptsize}}, requesting identifiers for newly found vulnerabilities requires the information of vulnerability type, vendor, and the affected equipment. Additional information such as impact and discoverer is also encouraged. 
In our research, we regard these required and encouraged information as profiling artifacts. Other vulnerability artifacts like CVSS score, CVSS vector are considered as non-profiling artifacts.

\subsection{GNN-based Entity Alignment}

Most GNN-based entity alignment methods ~\citep{gcnalign,cgmualign,rrea} are subject to the following framework: (1) a GNN to learn node representations from graph structure and (2) a margin-based loss to rank the distance between entity pairs. 
The loss function

\vspace{-10pt}
\begin{small}
\[L = \mathop\sum\limits_{(i, j) \in \mathcal{P}}\mathop\sum\limits_{(i', j') \in \mathcal{P^-}}\max\left\{\mathrm {d}(h_{i}, h_{j})-\mathrm {d}(h_{i'}, h_{j'})+\gamma, 0\right\}\]
\end{small}
\vspace{-2pt}
aims at making equivalent entities $(i,j)$ close to each other while maximizing the distance between negative pairs $(i', j')$. 
Here $h_i$ is the embedding of entity $i$ updated by a GNN layer 
in the form of:
\[
h^{l+1}_{i}\leftarrow \sigma\left(\mathrm{Aggregate}\left[W^l \cdot h^l_{k}, \forall k \in ({i}\cup N_{i})\right]\right)
\]
where $N{i}$ is the set of neighboring nodes around node $i$, $W^l$ is the transformation matrix in layer $l$, and $\sigma$ is a non-linear activation function. 
Instead of $W^l$, the relation-aware GNN learns a specific transformation matrix $W_r^l$ for each relation $r$. 
GNN variants serve the purpose of \texttt{Aggregate} by different operations such as normalized mean pooling \citep{gcnalign} and weighted summation \citep{gat}. 
%
In this paper, we propose \textit{masked aggregation} and \textit{partitioned attention} in \texttt{Aggregate} operation to infuse security domain knowledge into the learning of entity embeddings.

\begin{table*}[!htbp]
\centering
\begin{footnotesize}
\caption{Common artifacts provided by security repositories. Profiling artifacts are in bold.}
\label{tb:artifact}
\setlength{\tabcolsep}{3.6pt}{
\begin{tabular}{c|c|c|c|c|c|c|c|c|c|c}
\toprule
\multirow{2}*{\diagbox{Source}{Artifact}} &\multirow{2}*{CVE} & \multicolumn{2}{c|}{CWE}   &\multicolumn{2}{c|} {CVSS v2\&v3}   &\multicolumn{3}{c|} {CPE}
 & \multirow{2}*{\textbf{Impact}}  & \multirow{2}*{\textbf{Discoverer}}\\
&&\textbf{weakness} &cwe-id&vector&score & \textbf{vendor} & \textbf{product} &affected version&&\\
\midrule
ICS-CERT&$\oplus$&$\oplus$&$\oplus$&$\oplus$&$\oplus$&\checkmark&\checkmark & \checkmark&\checkmark&$\oplus$\\
\midrule
SF&\checkmark&$\ominus$&&&&\checkmark&\checkmark&\checkmark&\checkmark&\checkmark\\
\midrule

NVD&\checkmark&\checkmark&\checkmark&$\oplus$&$\oplus$&\checkmark&\checkmark&\checkmark&\checkmark&\\
\bottomrule
\end{tabular}}
\begin{scriptsize}\\$\dag$ (\checkmark: all reports, $\oplus$: large fraction of reports, $\ominus$: small fraction of reports). \end{scriptsize}
\end{footnotesize}
\end{table*}

\section{Security Knowledge Graphs}
In our study, we \textit{first time} construct and release three annotated security KGs from 
one security information portal, i.e., SecurityFocus\ignore{\footnote{https://www.securityfocus.com}} (SF)
and two governmental repositories, i.e, National Vulnerability Database\ignore{\footnote{https://nvd.nist.gov}} (NVD)
and ICS-CERT Advisories\ignore{\footnote{https://www.us-cert.gov/ics/advisories}} (CERT). 
All reports are in English. The crawled data are for informational purposes only, following all repositories' terms of service. Across the three KGs, we build two EA datasets by linking entities from ICS-CERT and SecurityFocus to NVD.
Below we explain the annotation process of the security KGs (Section \ref{sec:schema}, \ref{sec:ie}) and the EA dataset (Section \ref{sec:aligndata}). A quantitative study in Section \ref{sec:measurement} showing the particularity of the data demonstrates the challenges of aligning security entities. 
\subsection{KG schema}
\label{sec:schema}

\begin{figure}
    \centering
    \includegraphics[width=0.8\linewidth]{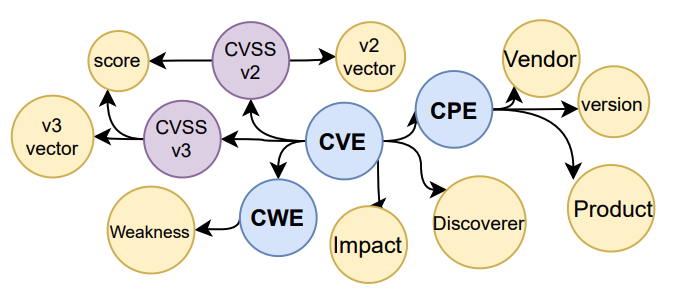}
    \begin{small}
    \vspace{-10pt}
    \caption{KG schema. Blue: entities; Yellow: literal artifacts; Purple: intermediate nodes. }
    \label{fig:BN}\end{small}
    \vspace{-10pt}
\end{figure}

Figure~\ref{fig:BN} illustrates a general schema (i.e., entity types and relations) of the proposed security KGs. 
We design the KG schema by summarizing the common artifacts\footnote{Examples and explanations are shown in Appendix \ref{appendix: example}.} provided by security reports, as shown in Table \ref{tb:artifact}.  These vulnerability artifacts can be interlinked by the concepts in the following standard security databases:
 Common Vulnerabilities and Exposures\ignore{\footnote{https://cve.mitre.org}} (CVE)
 for descriptions of publicly known vulnerabilities, Common Weakness Enumeration\ignore{\footnote{https://cwe.mitre.org}} (CWE)
 for categorizing software security weaknesses, Common Platform Enumeration\ignore{\footnote{https://nvd.nist.gov/products/cpe}} (CPE)
 for encoding names of IT products and platforms, and Common Vulnerability Scoring System (CVSS) for evaluating vulnerability severity.

%
\vspace{-5pt}
\subsection{KG Construction}
\label{sec:ie}
For NVD, we construct the security KG by retrieving vulnerability information from an NVD database~\citep{sepses} and linking them according to our KG schema.
For the other two semi-structured security repositories, we collected all available reports and annotated a subset of them. The annotation process are detailed as below.

\noindent\textbf{Gazeteer}.
We build a vocabulary for vulnerability-related artifacts based on CPE and CWE to help us recognize mentions of security entities in text. 
In total, the dictionary contains 26,461 product names, 8,738 vendors, and 187,711 versions from CPE, and 1,029 weaknesses from CWE.

\noindent\textbf{Annotation Process}. The entities and relations of security KGs are annotated by three cybersecurity-major students. The annotation standard is summarized through five preliminary rounds of annotation, covering 320 reports. 
In the first round, we discuss the context and the corresponding data entry of each vulnerability artifact in both repositories to summarize a set of patterns to extract entities and assign relations between them.
In each following round, we resolve disagreements and refine the patterns.
After the adjustment through five rounds, we programmatically parse all reports and manually verify a subset of them. 
The three annotators amend the annotation of each report in this subset, respectively, as shown in Table~\ref{tb:annotate}\footnote{To calculate Fleiss’s kappa, we take triples recognized by three annotators as samples, and the types of triples (including \textit{NotRecognized}) as categories.}. These annotations are merged into a final version by majority vote.
The statistics of three security KGs are shown in Table~\ref{tb:kgdata}. The data is released with the consent from all annotators.

\begin{table}
\centering
\begin{footnotesize}
\setlength{\tabcolsep}{2pt}{
\caption{Annotation statistics. The agreement is measured with Fleiss’s kappa coefficient \citep{fleiss1971measuring}.}

\label{tb:annotate}
\vspace{-10pt}
\begin{tabular}{ c|ccc } 
\toprule
\textbf{Repository} &{\# Reports} &{\# AnnotatedReports}&{Agreement}\\

 \midrule
ICS-CERT &1,324 &1,000  & 0.824 \\
 \midrule
 SecurityFocus&68,785 &  3,120& 0.865 \\
\bottomrule
\end{tabular}}
\end{footnotesize}
\end{table}

\begin{table}
\centering
\begin{footnotesize}
\setlength{\tabcolsep}{3pt}{
\caption{Statistics of security KGs. \#Ntypes represents the number of node types.}
\label{tb:kgdata}
\vspace{-10pt}
\begin{tabular}{ c|cccc } 
\toprule
\textbf{Security KG} &{\# Nodes} &{\# Edges}&{\# NTypes}&{\# Relations}\\
\midrule
NVD &24,069 &82,338  &11 &13 \\
 \midrule
ICS-CERT & 4,700&21,996  &12 & 14\\
 \midrule
 SecurityFocus& 7,011& 23,712 &7 & 8\\
\bottomrule
\end{tabular}}
\end{footnotesize}
\end{table}

\subsection{Annotation of Alignment Data}
\label{sec:aligndata}

We build two alignment datasets using the same data format with \citep{cgmualign}, which consists of two KGs along with positive and negative entity pairs. We consider NVD knowledge graph as the target KG and align the two other KGs to it, for NVD is the standard database with the highest coverage of vulnerabilities. 
Each alignment dataset includes vulnerability artifacts that are \textit{jointly} presented by both two KGs, except for the vulnerability identifiers, i.e., CVE-ID.
%
%
Records with two and above missing artifacts are dropped in this process.
Finally, we have 7,368, 1,136, and 2,410 vulnerabilities in KGs of NVD, ICS-CERT, and SecurityFocus for security entity alignment. 
We label the identical entities in different sources based on the same reference of common identifiers (e.g., CVE-ID, CWE-ID, CPE serial number) and security checklist references (i.e., hyperlinks of the same vulnerability to the other repositories) in the NVD database.
Table \ref{tb:eadata} shows the statistics of KGs to be aligned, including the number of co-occurred vulnerabilities and artifacts across two KGs and the total number of artifacts.
For negative alignments, we randomly sample 10 negative entities from the other KG for each entity in all positive pairs. We also sample 10 negative pairs for entities on the source KG without a corresponding alignment on the target KG. In total, the alignment dataset contains 2,594 positive pairs and 65,710 negative pairs.

\begin{table}
\centering
\begin{footnotesize}
\setlength{\tabcolsep}{6.2pt}{
\caption{Statistics of KGs for Alignment.}
\label{tb:eadata}
\begin{tabular}{ c|cc c} 
\toprule
\textbf{Dataset}&{\#AlignedEnt} &{\#SharedArt} & \#TotalArt\\
\midrule

 CERT-NVD &987 &514&19,721\\
 SF-NVD &1,607 &1,303&20,000\\
\bottomrule
\end{tabular}}
\end{footnotesize}
\end{table}

\subsection{Findings}
\label{sec:measurement}
In the security KGs, we observe the attribute inconsistency between
positive entity pairs and attribute similarity between negative entity pairs, compared with entity alignment datasets in other domains. We elaborate our findings as below.

\noindent\textbf{Inconsistencies in positive pairs}. 
We denote that a pair of entities encounter an attribute inconsistency in a certain type if no match exists between their attributes in this type.
Here, a match means the tokens are the same, or one token is a sub-string of another.
Among all the positive pairs in our dataset, 56.0\% pairs of vulnerabilities have attribute inconsistency in more than half of attribute types.
%
However, in the entity alignment datasets of music and movie \citep{cgmualign}, only 8.3\% positive entity pairs face the same problem. 
The attribute inconsistencies hinder the classifier from making correct positive decisions. 
To alleviate this problem, we propose a masked gate to aggregate neighboring attributes selectively.

\noindent\textbf{Similarity in negative pairs}. 
We observe that different vulnerabilities can be associated with quite many identical artifacts.
More specifically, among all negative pairs in our alignment datasets, 4.04\% differ in only one quarter types of artifacts.
As a comparison, with the same negative sampling ratio, only 1.14\% negative pairs in the music dataset in \citep{cgmualign} have the same problem.
Such circumstances make it difficult for the classifier to distinguish different-but-similar entities.
Based on the vulnerability profiling standard and our observation on the datasets, to distinguish a vulnerability from a set of known vulnerabilities, at least one of its \textit{profiling} artifacts should be different.
Therefore, higher attention on the profiling artifacts helps preserve the distinction between similar entities. Accordingly, we propose a partitioned attention mechanism to assign higher importance weights to the profiling artifacts in a mandatory way.

\vspace{-2pt}
\section{Problem Formulation}
\vspace{-5pt}
We represent the security knowledge graph as a heterogeneous graph $G$ connecting nodes in types $\mathcal{A}$ by relations $\mathcal{R}$.   
We use a set of nodes  $\mathcal{V}$ to denote security entities with a mapping $\psi:\mathcal{V}\rightarrow\mathcal{A}$ to the entity types. The security KG is modeled as a set of triples $G(\mathcal{V}, \mathcal{A}, \mathcal{R})=\{t|t:(i, r, j), i, j\in \mathcal{V}, r\in\mathcal{R}\}$. 
Entity types and relations are aligned in advance:
for a certain $G^*$, its entity types $\mathcal{A^*}$ and relations $\mathcal{R^*}$ are aligned to the KG schema in Section \ref{sec:schema} during KG construction.

\noindent\textbf{Security entity alignment.}
Given a source knowledge graph ${G}(\mathcal{V}, \mathcal{A}, \mathcal{R})$ and a target knowledge graph $G'(\mathcal{V'}, \mathcal{A'}, \mathcal{R'})$, entity alignment aims at finding the identical target entity $v'\in\mathcal{V'}$ for each source entity $v\in\mathcal{V}$.
However, in the security domain, only a few entities on one KG has the same real-world reference on another.
Therefore, we define the security entity alignment problem as follows:
Given a set of entity pairs $S=\{(v_s, v_s')\in \mathcal{V}_s\times \mathcal{V'}_s, \mathcal{V}_s \subset\mathcal{V},\mathcal{V'}_s \subset \mathcal{V'}\}$ between two knowledge graphs $G$ and $G'$, security entity alignment aims to determine whether each entity pair in $\{(v_i, v_i')\in (\mathcal{V} \times \mathcal{V'}) \setminus S\}$ refer to the same real word objects with high precision and recall.
\vspace{-2pt}
\section{Method}
\vspace{-5pt}
CEAM applies a joint GNN framework to address the security entity alignment problem. 
CEAM consists of an aggregate layer that initializes entity representation by masked neighborhood information, a 2-layer GNN that embeds structural information in node representation,  and a classifier that makes the alignment decision according to node distance. As in~\citep{cgmualign}, we model the security entity alignment as a binary classification problem on a given set of positive and negative entity pairs. 
The entire model is jointly trained by the cross entropy loss:
$L =- \frac{1}{N}\sum_i[y_i\log p_i+(1-y_i)\log(1-p_i)]$, where $y_i$ labels whether the two entities in pair $i$ refer to the same object.
Below we describe the two proposed mechanisms that enhance the GNN to learn better embeddings for security entities. 

\vspace{-5pt}
\subsection{Masked Attribute Aggregation}
\vspace{-5pt}
\label{sec:mask}

We use masked attribute aggregation to represent ID-like entities from neighborhood literal artifacts.
Due to the heterogeneity of knowledge graphs, we first aggregate the representation of artifacts in the same type into relation representation. Further, we use the similarity between the relations representation of different entities to calculate the mask, an attribute-level scaling operator for aggregating relation representation into entity representation. The mask relaxes the similarity constraint of a candidate pair of entities if they are mutually the most closed ones over the two KGs.
%
Below we elaborate on the two-stage masked aggregation.
%
%

%
\noindent\textbf{Stage 1: towards relation representation}.
For node $i$ on the graph $G(\mathcal{V}, \mathcal{A}, \mathcal{R})$, 
its corresponding relation $r$ is represented as

\vspace{-5pt}
\begin{small}
\[
    \phi_i(r) = \sum_{j\in N_{i,r}}{\alpha_{irj}  W_rh_j},
\]
\end{small}
where $W_r$ is the transform weight for relation $r$, $N_{i,r}=\{j|(i, r, j)\in G\}$ represents the neighbors associated with $i$ by relation $r$,
 $\alpha_{irj}$ characterizing the importance of node $j$ to node $i$ is the relative co-occurrence rate of two nodes across two KGs:

\vspace{-5pt}
\begin{small}
\begin{equation*}
     \alpha_{irj} =  \frac{\exp{\left(-d_j(d_j+d_j')^{-1}\right)}}{\sum\limits_{k\in N_{i,r}}\exp{\left(-d_k(d_k+d_k')^{-1}\right)}},
\end{equation*}
\end{small}
where $d_j$ and $d_j'$ are the number of entities connected to node $j$ on graph $G$ and $G'$, respectively.
Such design follows two considerations: 1) the neighbors connected with fewer entities on the source side are more specific and representative to the entity, and 2) the neighbors that are jointly linked with the positive entity pairs will benefit the alignment.  
For example, NVD provides both devices and configures (e.g., software, hardware, etc.) as the affected products of a cyber vulnerability, while most reports in CERT mainly contain device information. 
Therefore, it is more likely to boost a positive align if the attributes of the device are emphasized to represent the relation \texttt{affected products}.  
Also, some basic architectures in cyber systems, like SCADA (Supervisory ControI And Data Acquisition System), which frequently appear in various security reports, are less representative to profile a specific vulnerability or distinguish different vulnerabilities. 

\noindent\textbf{Stage 2: towards entity representation}.
By stacking the relation representation $\phi_i(r)$, entity $i$ is temporarily encoded by the matrix: $\Phi_i = \left(\phi_i(r_1)^T, \phi_i(r_2)^T, ..., \phi_i(r_{|\mathcal{R}|})^T \right)$.
Further, we study a mask gate that scales the attributes (i.e., features) of the relation representation during the aggregation.
To do this, we consider a scaling parameter $m$ for each attribute $t$ in the representation of $r$, which reflects the confidence that entity $i$ possesses the attribute. 
We construct a set of candidates $\mathcal{C}_{i}=\{i'|\exists j \in (\mathcal{V}\cap\mathcal{V'}): (i,r,j)\in\mathcal{G}\wedge (i',r,j)\in\mathcal{G}'\}$ for entity $i$.
The scaling parameter $m_{i,r}^t$ is measured by the consistency of the attribute $t$ in the representation of $r$ among all candidate pairs $\{(i, i')\}$ of $i$, 
\[m_{i,r}^t=\exp(-\sum_{i' \in\mathcal{C}_{i}}c_{ii'}[\phi_i^t(r)-\phi_{i'}^t(r)]^2),
\]
where $\phi_i^t(r)$ is the value of attribute,  $c_{ii'} = \mathop \mathrm{softmax}_{i'\in \mathcal{C}_i}\left(-||\Phi_i-\Phi_{i'}||_F\right)$ indicates the correspondence of $i'$ to $i$.
Therefore, for the complete relation representation $\phi_i(r)$, we can define the mask gate altogether as a \textit{diagonal} matrix $M^r_i$,

\vspace{-15pt}
\begin{small}
\[
    M^r_i = \exp\left(-\sum_{i'}c_{ii'}\cdot \mathbf{d}_r(i, i')\mathbf{d}_r(i, i')^T\right),
\]
\end{small}
where $\mathbf{d}_r(i, i')= \phi_i(r)-\phi_{i'}(r)$.
Finally we have the aggregated representation of entity $i$ as:
\[
    h_i = \frac{1}{|\mathcal{R}|}\sum_r{M^r_i\phi_i(r)}
\]
 The mask reduces the effect of different attributes between a candidate pair of entities only if they are \textit{mutually} the most closed ones to each other among two KGs; Otherwise, the distance between the masked entity representations will not decrease since the masks are asymmetric and scale different attributes for the two entities. This enables CEAM to achieve high recall while preserving tolerable precision for security entity alignment.

\subsection{GNN with Partitioned Attention}
\label{sec:attn}
In GNN encoder, we learn transformations $W_t$ for each entity type and $W_r$ for each relation. As in  ~\citep{cgmualign}, we calculate node-level attention $s$ and relation-level attention $\beta$ to accumulate neighborhood information:

\vspace{-5pt}
\[
    z_i = \sum_{r,j\in N_{i,r}}{s_{irj}\cdot\beta_{ir}\cdot W_rh_j},
\]
\noindent and combines the self information $h_i^{(l-1)}$ and accumulated neighborhood information $z_i$: 
\begin{equation*}
    h_i^{(l)} = \sigma\left([W_{t}^{(l)}\cdot h_i^{(l-1)}||z_i]\right).
\end{equation*}
The node-level attention $s_{irj}$ is defined as the cosine similarity between two connected nodes:
\begin{equation*}
       s_{irj} = \mathop\mathrm{cos\_sim}<W_{t}^{(l)} h_i^{(l-1)}, W_r^{(l)} h_j^{(l-1)}>,
\end{equation*}
and the relation-level attention $\beta_{ir}$ is collectively decided by whether the relation is predefined as critical and the similarity between nodes.
We consider the profiling relations $R_P$ as the set of relations  connecting profiling artifacts (See Section \ref{sec:background-vul}) with entities. 
Let the profiling ratio $\rho = \frac{|R_p|}{|R|}$. Accordingly, we define the partisan term $\delta$ with a positive hyper-parameter $\epsilon$ where $0<\epsilon<1-\rho$:

\vspace{-10pt}
\begin{small}
\begin{equation*}
\delta_{R_X} = \left\{\begin{array}{rcl}\rho+\epsilon-0.5&if&R_X=R_P\\ -\rho -\epsilon+0.5&if&R_X=R\setminus R_P\end{array}\right.,
\end{equation*}
\end{small}
where $R_X$ represents the group of profiling or non-profiling relations.
And the attention weight for $r\in R_X$ is

\vspace{-10pt}
\begin{small}
\begin{equation*} 
    \beta_{ir} = (\frac{1}{2}+\delta_{R_X} )\mathop\mathrm{softmax}\limits_{r\in R_X}\left(|N_{i,r}|^{-1}\sum_{j\in N_{i,r}}s_{irj}\right),  
\end{equation*}
\end{small}
Such a mechanism ensures that critical relations (e.g., affected products, etc.) suggested by vulnerability profiling standards are emphasized while reserving the relative importance within a relation group (i.e., profiling or non-profiling). 

\section{Experiments}
\begin{table*}
\centering
\begin{small}
\setlength{\tabcolsep}{7pt}{
\caption{Results of Cybersecurity Entity Alignment.}
\label{tb:eacompare}
\vspace{-10pt}
\begin{tabular}{ c|c|c|c|c|c|c} 
\toprule
\multirow{2}*{\textbf{Method}} & \multicolumn{3}{c|}{\textbf{CERT-NVD}} &\multicolumn{3}{c}{\textbf{SF-NVD}} \\
&Pre@Rec=.95 & F1 & PRAUC&Pre@Rec=.95 & F1 & PRAUC\\
\midrule
GCN&0.198 $\pm$ .002 &     0.330 $\pm$ .004 &   0.265 $\pm$ .006  &   0.139 $\pm$ .004 &   0.233 $\pm$ .003 & 0.172 $\pm$ .002\\

GAT &      0.201 $\pm$  .005 &     0.340 $\pm$ .008 &   0.357 $\pm$ .003  &     0.165 $\pm$ .002 &   0.274 $\pm$ .001 & 0.293 $\pm$ .002\\
GraphSage &0.179 $\pm$ .001& 0.327 $\pm$ .003 & 0.303 $\pm$ .004 & 0.128 $\pm$ .005 &   0.216 $\pm$ .003 & 0.207 $\pm$ .001\\ 
R-GCN  &
    0.361$ \pm$ .003 &     0.475 $\pm$ .003 &   0.421 $\pm$ .005  & 0.267 $\pm$ .002 &   0.456 $\pm$ .003 & 0.412 $\pm$ .002\\

R-GAT  & 0.392 $\pm$ .004 & 0.610 $\pm$ .002 & 0.654 $\pm$ .003 & 0.293 $\pm$ .004& 0.541 $\pm$ .006& 0.562 $\pm$ .004\\
R-GraphSage&0.290 $\pm$ .002 & 0.450 $\pm$ .005& 0.413 $\pm$ .009&0.199 $\pm$ .005& 0.414 $\pm$ .018& 0.382 $\pm$ .010\\
\midrule
PARIS& -.- & 0.712 $\pm$ .000& 0.543 $\pm$ .000 & -.- & 0.611 $\pm$ .000& 0.346 $\pm$ .000\\
PRASE(BootEA) &  -.- & 0.731 $\pm$ .000& 0.576 $\pm$ .000 & -.-&  0.620 $\pm$ .000 & 0.477  $\pm$ .000\\
PRASE(MultiKE)&  -.- & 0.706 $\pm$ .000& 0.512 $\pm$ .000 & -.-& 0.609 $\pm$ .000& 0.398 $\pm$ .000\\

CGMuAlign & 0.470 $\pm$ .026 & 0.761 $\pm$ .012 & 0.720 $\pm$ .003 & 0.328 $\pm$ .013& 0.654 $\pm$ .021& 0.576 $\pm$ .007\\
 \midrule
CEAM& 0.778 $\pm$ .015 & 0.843 $\pm$ .015   &0.875 $\pm$ .002& 
0.694 $\pm$ .011 &  0.787 $\pm$ .021 & 0.801 $\pm$ .004\\

\bottomrule
\end{tabular}}
\end{small}
\end{table*}

\subsection{Experimental Settings}

\noindent\textbf{Dataset and data  processing}. We set the train-test-split rate as 0.25, and use a 5-fold cross validation on the training set to select hyperparameters, e.g., the learning rate.
For the alignment targets, i.e., vulnerability entities, we replace the original CVE-IDs with randomly assigned numerical IDs so entities in a positive pair are represented by different tokens. 
Other artifact entities are initialized with textual representations.
%
We train a Word2Vec~\citep{church2017word2vec} model using Gensim \citep{vrehuuvrek2011gensim} with all available reports, and use fastText~\citep{joulin2016fasttext} to encode textual features of artifact entities as 100-dimensional vectors for all of the GNN-based methods.

\noindent\textbf{Evaluation metrics}.
We apply three metrics: Precision@Recall=0.95, F1, and PRAUC (precision-recall area under curve).
%
For the first metric, we tune the alignment threshold during testing and report the highest precision when the recall reaches the given value. 
A higher recall value is more desirable in the security applications \citep{yang2021few} to recover the identities of vulnerabilities missing identifiers, where the false positives can be excluded through handful manual verification. 
For F1 evaluation we use the decision threshold that achieves the highest macro F1 during validation.

\noindent\textbf{Hyperparameters}. 
For each model, we apply a binary search from 0.001 to 0.1 on the validation set for learning rate selection. The best learning rate for CERT-NVD dataset and SF-NVD dataset are 0.02 and 0.025, respectively. The dimension of GNN embeddings in CEAM and baselines is 64.

\noindent\textbf{Computation Cost}. The complexity of CEAM is $O(|\mathcal{V}|\cdot (N+ C) \cdot \frac{S}{B})$, where $N$ is the maximum neighborhood size, $C$ is the maximum candidate size, $S$ is the training size and $B$ is the batch size. The model has 340,309 parameters. We run CEAM on Intel Core i5-3550 3.30GHz 4-Core Processor with NVIDIA GTX 1070 Ti GPU. Averagely, each batch of size 128 costs 295.3 ms for training. 
%
%
\vspace{-5pt}
\subsection{Baselines}
\vspace{-2pt}
\noindent\textbf{EA w./o. one-to-one constraint.} We compare the proposed method with the following state-of-the art methods:
\textit{CG-MuAlign}  \citep{cgmualign} features a collective GNN framework to align entities in two domains: music and movie. It also formalizes the alignment problem as a classification task on a given set of entity pairs. \textit{PARIS} \citep{paris} is a conventional method that align entities by probabilistic reasoning. Several studies \citep{seu,zhao2020experimental} have noted that PARIS outperforms many ``advanced'' EA models. \textit{PRASE} ~\citep{prase} is an 
extension of PARIS, which incorporates the translation-based EA models to boost the alignment decisions.







\noindent\textbf{GNN variants}.
We evaluate the recent GNN models in security entity alignment, as shown in Table \ref{tb:eacompare}. 
R-GCN, R-GAT, and R-GraphSage are the variants of GCN, GAT, and GraphSage which train specific transformation weights for each relation.
%
We reimplement all of the GNN variants with DGL \citep{wang2019dgl}. CEAM and all the GNN variants use the same model structures: 2-layer GNNs following with two fully-connected layers. 

We use the EA datasets in Section \ref{sec:aligndata} to evaluate CEAM, CGMuAlign, and all GNN variants. For PARIS and PRASE, we generate datasets with the format in \citep{prase}.  
To ensure the comparability, for the results of PARIS and PRASE, we only consider the wrong alignments sampled in our negative pairs (See Section \ref{sec:aligndata}) as false positives.
\vspace{-3pt}
\subsection{Effectiveness}
\vspace{-2pt}
Table \ref{tb:eacompare} shows the averaged alignment results of three runs on the two datasets. 
All models show higher performance on CERT-NVD dataset than SF-NVD dataset. The main reason is that SF knowledge graphs provide fewer entity types and suffer more from sparsity issues. Overall, CEAM outperforms all other methods. 
CG-MuAlign, which is similar in problem formulation and model architecture with CEAM, shows the second best result. 
PARIS is less successful to identify the same entities in security KGs. Compared with PARIS, PRASE improves 1.9\% on averaged F1 by incorporating translation-based alignment modules.
%
Note that CEAM has significantly better performance on Pre@Rec=0.95. This is mainly because the proposed mechanisms enable CEAM to achieve high recall even with a rather large decision threshold. In contrast, previous EA models attach less importance to preventing false negatives and the decision threshold needs be very small to achieve high recall, which greatly reduces the precision.
More specifically, CEAM employs masked aggregation to relax the similarity constraints for positive aligns, which decreases false negatives to achieve high recall. In the meantime, such relaxation is only applied when the two entities are mutually the most closed candidate to each other, which preserves evidence for negative aligns to prevent low precision. 
\vspace{-5pt}
\subsection{Ablation Study}
Table~\ref{tb:analysis} shows the improvement by each proposed mechanism, respectively. All the scores are averaged on the two alignment datasets.
%
The variant without both two mechanisms is equivalent to the baseline model using R-GAT to train node representations. 
Overall, CEAM outperforms baselines by 23\% F1 score. Below we detail the quantative improvement by the two proposed mechanisms.

\begin{table}
\centering
\begin{footnotesize}
\setlength{\tabcolsep}{3pt}{
\caption{Analysis of proposed method.}
\label{tb:analysis}
\vspace{-10pt}
\begin{tabular}{ c|ccc } 
\toprule
\textbf{Method} &{P@R=0.95} &{F1}\\
\midrule
\textbf{w.o.} masked aggregation &0.516&0.712 \\

 \textbf{w.o.} mask (=mean aggregate)&0.532 &0.763  \\
 \midrule
\textbf{w.o.} partitioned attention &0.687 & 0.790  \\
\textbf{w.o.} non-profiling artifacts &0.665 &0.775  \\
\midrule
CEAM &0.736 &  0.815 \\
\bottomrule
\end{tabular}}
\end{footnotesize}
\end{table}

\noindent\textbf{Masked aggregation}.
The first row of Table~\ref{tb:analysis} shows the model effectiveness without any aggregation. 
The second row shows the result where the vulnerability entity representations are initialized as the mean of its neighboring artifacts.
The results show that the aggregation operation boost the performance by 5.1\% in averaged F1, and the mask mechanism further improves 5.2\% averaged F1. 

\noindent\textbf{Partitioned attention}.
We present the alignment result using traditional attention mechanism where $    \alpha_{irj}=\mathop\mathrm{softmax}\limits_{r\in R, j\in N_{i,r}}\sigma\left(\overrightarrow{w_r}^T\left[h_i||W_rh_j\right]\right)$ and $z_i=\sum_{r\in R}\sum_{j\in N_{i,r}}\alpha_{irj}W_rh_j$ in the third and fourth row of Table~\ref{tb:analysis}. 
Compared with traditional attention operation on all entity types, the partition mechanism enhances 2.5\% averaged F1, since the profiling artifacts are enforced to be emphasized to alleviate the noises introduced by non-profiling artifacts.
We also compare the result on dataset that only employs profiling artifacts while dropping the non-profiling ones. The result shows that the non-profiling artifacts still provide useful information to improve 1.5\% F1. 
We evaluate the performance varying the hyper-parameter $\epsilon$ (See Section 5.2) and the result is shown in Figure \ref{fig:eps} in Appendix \ref{sec:eps}.


\section{Discussion}
\vspace{-5pt}
\noindent\textbf{Findings}.
Among the 3,546 annotated vulnerabilities, 52 are released in reports without the identifier. 
Such vulnerability entities do not appear in the entity alignment datasets since their corresponding NVD vulnerabilities are unkown.
We construct the candidate pairs for such vulnerabilities as described in Section \ref{sec:mask} and run CEAM to predict their binary labels.
Finally, we manually verify the \textit{positive} alignments according to the reference checklist and the artifacts provided from both sides, and find 70.2\% of them are correct, which recovers the missing vulnerability identifiers for 28 reports.

\noindent\textbf{Limitations}.
Our research demonstrates that the proposed domain-driven mechanisms benefit security entity alignment. 
Most false positives are produced in vulnerabilities with similar impacts and the same other artifacts.
The representations of such underlying different vulnerabilities are very closed, and it's hard to distinguish them for all similarity-based alignment methods.
Meanwhile, the false negatives are mainly due to data quality issues such as information inconsistency, which is commonly observed in real-world vulnerability repositories \citep{dong2019towards,anwar2021cleaning,jiang2021evaluating}. For example, the affected product, vendor, and weakness of CVE-2018-7084 provided by ICS-CERT and NVD are all different. Although the mask mechanism relaxes the similarity constraint, CEAM cannot align the two records in this case.
We leave the entity alignment with information quality assessment as future work.
%
%

\section{Related Works}
\vspace{-5pt}
\noindent\textbf{Graph Entity Alignment}.
Recent graph entity alignment methods assume that same entities on different KGs have: (1)
similar attribute distributions, and (2) similar neighborhood structures. 
Accordingly, translation-based methods~\citep{hao2016joint,chen2016multilingual,zhu2017iterative,sun2017cross,sun2018bootstrapping,hu2019multike} use knowledge graph 
embedding models to generate entity embeddings from triple structures; GNN-based methods~\citep{gcnalign,wu2019jointly,mugnn,wu2020neighborhood, emgcn, rrea,zhu2021raga} apply graph convolution operations to utilize neighborhood information. Further, \citep{pei2020rea} proposes reinforced training strategy to achieve noise-aware entity alignment; \citep{yan2021dynamic} proposes topology-invariant gates to dynamically align evolving knowledge graphs.
By applying graph isomorphism, \citep{seu} learns a permutation matrix that transforms one KG to another by reordering the entity node indices.

\noindent\textbf{Domain-specific entity linking}. 
To disambiguate candidate entities for the given mentions, \citep{DBLP:conf/aclnews/InanD18} utilizes domain information to filter candidates;
~\citep{DBLP:journals/tkde/ShenHWYY18} designs a probabilistic
linking model that combines the distribution of entities and domains;
~\citep{DBLP:conf/acl/KlieCG20} proposes entity linking for low-source domains by letting human annotators make corrections on ranked candidate entities for each mention.
All existing works utilize domain information to constrain the search space of candidates, while the domain-specific properties are overlooked.

\noindent\textbf{Security entity identification}. Cross-platform entity identification has been studied in the security domain to combat the evolving cybercrimes.
\citep{xu2017neural} detects the similarity of cross-platform binary code functions based on a graph embedding generation network.
\citep{zhang2019key} and  \citep{fan2019idev} build weighted multi-view networks based on user relatedness and apply GCN for node representations to identify key players in cybercriminal communities or link cross-platform suspicious accounts. 
To the best of our knowledge, we are the first to incorporate the properties of security entities into the alignment process.


\vspace{-5
pt}
\section{Conclusion}
\vspace{-10pt}
 In this paper, we construct the first annotated cybersecurity-domain entity alignment datasets and reveal the attribute inconsistency feature of security entities. Based on this feature, we propose the first cybersecurity entity alignment model, CEAM, which equips GNN-based entity alignment model with two mechanisms, asymmetric masked aggregation and partitioned attention. Experimental results on cybersecurity-domain entity alignment datasets demonstrate the effectiveness of our method, which outperforms the state-of-the-art.

\bibliography{ref}
\bibliographystyle{acl_natbib}

\appendix
\section{Entities and Relations}
\label{appendix: example}
We present concrete examples of vulnerability artifacts in Table \ref{tb:artifact} and the explanations as below. \textit{CVE} is the common identifiers of cybersecurity vulnerabilities. \textit{Weakness} characterizes the category of the vulnerability, and \textit{CWE\_ID} is the identifier of Weakness. \textit{Product} and \textit{Vendor} are the name and the provider of the affected products (e.g., software, hardware, device, etc). \textit{Version} is short for affected versions. \textit{Impact} is the consequence of exploiting the vulnerability. \textit{Discoverer} is the name of the person or the organization who reported the vulnerability. In a triplet $(h,r,t)$, the relation $r$ is decided by the types of the head entity $h$ and the tail entity $t$ and is in the form of \textit{hasTail}. For example, the relation between a \textit{Vulnerability} entity and a \textit{Discoverer} is \textit{hasDiscoverer}.
\begin{table}
\centering
\begin{small}
\caption{Examples of vulnerability artifacts. } 
\label{tb:artifact}
\begin{tabular}{|c|c|}
\hline
\textbf{Artifact} & \textbf{Example}\\
\hline
CVE & CVE-2019-34623\\
\hline
CWE\_ID & CWE-200\\
\hline
Weakness & Information Exposure\\
\hline
Vendor& Siemens\\
\hline
Product& SINEMA Remote Connect Server \\
\hline
version&   2.0 SP1\\
\hline
Impact &denial of service\\
\hline
Discoverer & Hendrik Derre, Tijl Deneut\\
\hline
CVSSv3 vector& AV:N/AC:L/PR:L/UI:N/S:U/C:L/I:N/A:N\\
\hline
CVSSv2 vector& AV:N/AC:L/Au:S/C:P/I:N/A:N\\
\hline
CVSSv3 score& V3 4.3\\
\hline
CVSSv2 score& V2 4.0\\
\hline
\end{tabular}
\end{small}
\end{table}

\ignore{
\subsection{Patterns for structured text information extraction}
\label{appendix: }
\begin{table}[h!]
\centering\small
\caption{Patterns for structured text information extraction. }
\label{tb:rules}
\begin{tabular}{cc}
\toprule
Entity Class & Pattern \\ 
\midrule
Vuln Type &  Vulnerability:[ A-Za-z]+\\
\hline
\multirow{2}{*}{Version} & ([Aa]+ll )?[Vv]+e?r?(sion)?s?( (prior to )?\\
 & ($\backslash$w?$\backslash$d+.*)+( and ((prio$|$earlie)r)$|$before)?)?\\
\hline
CVE-ID & CVE-[0-9]\{4\}-[0-9]\{4,\}\\
\hline
CWE-ID & CWE-[0-9]\{1,\}\\
\hline
CVSS Vector & ([A-Za-z]+:[A-Za-z]+/?)\{2,\} \\
\hline
CVSS Score & (base$|$temporal) score of [0-9]+.[0-9]+\\
\hline
Metric Value & Attention:[ A-Za-z]+\\
\hline
\multirow{2}{*}{Intrusion} & (result(s$|$ing)? in|causes?$|$allows?( attakers?)?)\\
&[A-Za-z ]+(,$|$.)\\
\hline
Product & Equipment:[A-Za-z0-9]+.?[A-Za-z0-9]*\\
\hline
Vendor & Vendor:[A-Za-z0-9]+\\
\hline
Discoverer & [A-Za-z ]+(releas$|$discover$|$report)ed \\
\bottomrule
\end{tabular}
\end{table}}

\section{Hyper-parameter Analysis}
\label{sec:eps}
\begin{figure}
    \centering
    \includegraphics[width=1\linewidth]{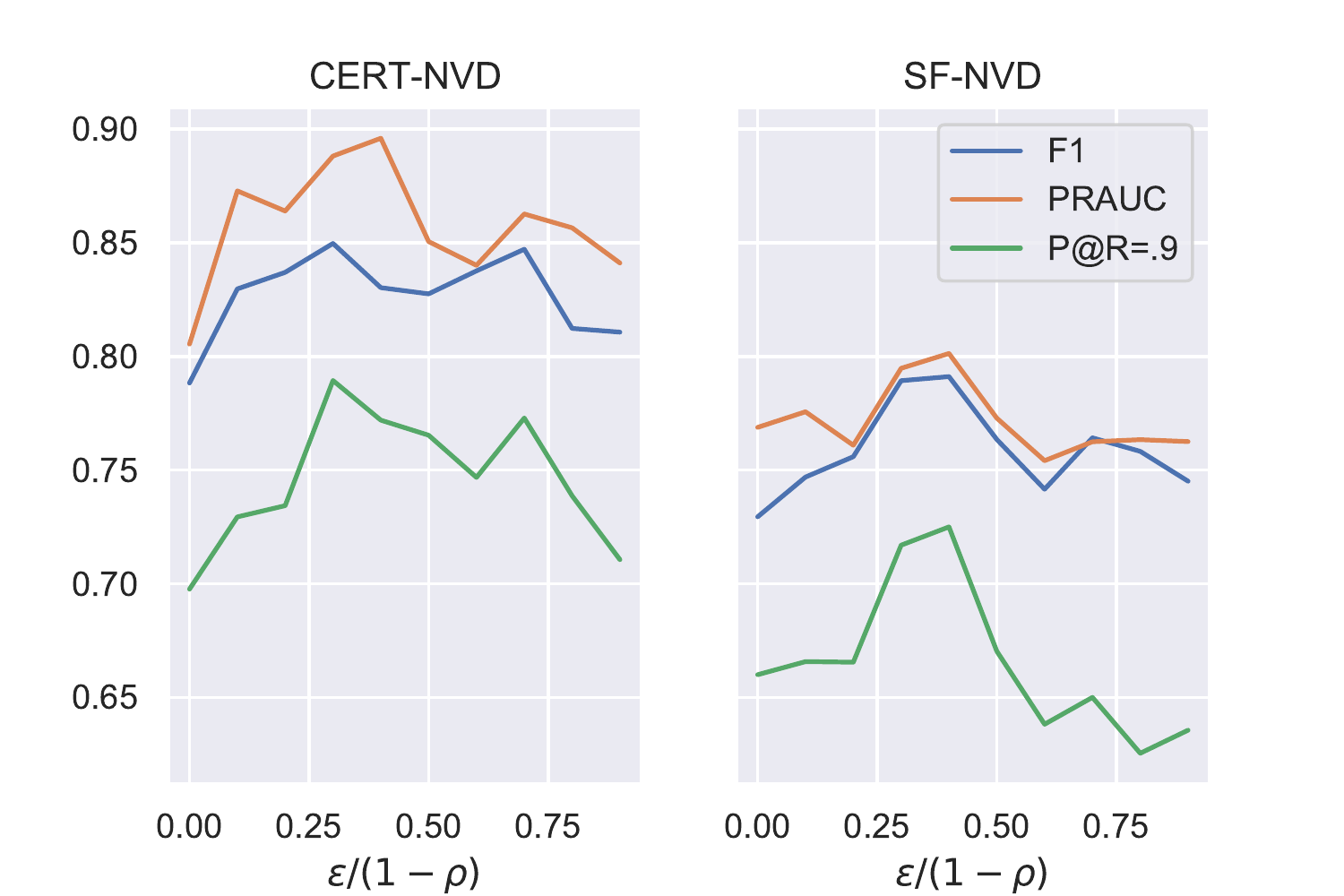}
    \begin{small}
    \caption{Model performance with different partisan term. X-axis shows the ratio between $\epsilon$ and $1-\rho$.}
    \label{fig:eps}\end{small}
\end{figure}

We evaluate the performance varying the partisan term decided by $\epsilon$ in Section 5.2 and 
the results are shown in Figure \ref{fig:eps}. The profiling ratio $\rho$ (i.e., the portion of profiling relations among all relations) of the two alignment datasets (i.e., CERT-NVD and SF-NVD) are 0.4 and 0.8, respectively. 

\end{document}